%% file: main.tex
\documentclass[runningheads]{llncs}
\usepackage{graphicx}
\usepackage[utf8]{inputenc} 
\usepackage[T1]{fontenc}    
\usepackage{hyperref}       
\usepackage{url}            
\usepackage{booktabs}       
\usepackage{amsfonts}       
\usepackage{nicefrac}       
\usepackage{microtype}      

\usepackage{hyperref}

\usepackage{url}
\usepackage{algorithm}
\usepackage{algorithmic}
\usepackage{graphicx}
\usepackage{amssymb}
\usepackage{multicol}
\usepackage{amsmath}
\usepackage{array}

\newcolumntype{x}[1]{%
>{\centering\hspace{0pt}}p{#1}}%

\usepackage{float}
\begin{document}
\title{Unified Models of Human Behavioral Agents in Bandits, Contextual Bandits and RL}
\titlerunning{Human Behavioral Agents in Bandits, Contextual Bandits, and RL}

\input{./sec_author.tex}

\maketitle

\input{./sec_abstract.tex}
\input{./sec_intro.tex}
\input{./sec_problem.tex}
\input{./sec_method.tex}
\input{./sec_experiment.tex}
\input{./sec_conclusion.tex}

\bibliographystyle{splncs04}
\bibliography{main}

\input{./sec_appendix}

\end{document}

%% file: sec_author.tex

\author{Baihan Lin\inst{1}, Guillermo Cecchi\inst{2}, Djallel Bouneffouf\inst{2}, Jenna Reinen\inst{2}, Irina Rish\inst{3}}
%
\authorrunning{B. Lin et al.}
%
\institute{Center for Theoretical Neuroscience, Columbia University, New York, USA \and
IBM Thomas J. Watson Research Center, Yorktown Heights, NY, USA \and
Mila - Quebec AI Institute, University of Montreal, Montreal, Canada
\email{baihan.lin@columbia.edu, gcecchi@us.ibm.com, \{djallel.bouneffouf,jenna.reinen\}@ibm.com, irina.rish@mila.quebec }\\}






%% file: sec_abstract.tex
\begin{abstract}

Artificial behavioral agents are often evaluated based on their consistent behaviors and performance to take sequential actions in an environment to maximize some notion of cumulative reward. However, human decision making in real life usually involves different strategies and behavioral trajectories that lead to the same empirical outcome. Motivated by clinical literature of a wide range of neurological and psychiatric disorders, we propose here a more general and flexible parametric framework for sequential decision making that involves a two-stream reward processing mechanism. We demonstrated that this framework is flexible and unified enough to incorporate a family of problems spanning multi-armed bandits (MAB), contextual bandits (CB) and reinforcement learning (RL), which decompose the sequential decision making process in different levels. Inspired by the known reward processing abnormalities of many mental disorders, our clinically-inspired agents demonstrated interesting behavioral trajectories and comparable performance on simulated tasks with particular reward distributions, a real-world dataset capturing human decision-making in gambling tasks, and the PacMan game across different reward stationarities in a lifelong learning setting.\footnote{The codes to reproduce all the experimental results can be accessed at \url{https://github.com/doerlbh/mentalRL}.} 

\keywords{Reinforcement Learning \and Contextual Bandit  \and Neuroscience}
\end{abstract}

%% file: sec_intro.tex
\section{Introduction}

In real-life decision making, from deciding where to have lunch to finding an apartment when moving to a new city, and so on, people often face different level of information dependency. In the simplest case, you are given $N$ possible actions (``arms''), each associated with a fixed, unknown and independent reward probability distribution, and the goal is to trade between following a good action chosen previously (exploitation) and obtaining more information about the environment which can possibly lead to better actions in the future (exploration). The {\em multi-armed bandit (MAB)} (or simply, bandit) typically models this level of exploration-exploitation trade-off \cite{LR85,UCB}. In many scenarios, the best strategy may depend on a context from current environment, such that the goal is to learn the relationship between the context vectors and the rewards, in order to make better prediction which action to choose given the context, modeled as the {\em contextual bandits (CB)} \cite{AgrawalG13,11}, where the context can be attentive \cite{bouneffouf2017context,lin2018contextual}, budgeted \cite{lin2021optimal} or clustered \cite{lin2020semi,lin2020voiceid,lin2020speaker}. The learning can be purely online or pretrained in offline fashion \cite{lin2021offlinebandit}. In more complex environments, there is an additional dependency between contexts given the action an agent takes, and that is modeled as Markov decision process (MDP) in {\em reinforcement learning (RL)} problem \cite{Sutton1998}. 

To better model and understand human decision making behavior, scientists usually predict the trajectories \cite{lin2020predicting} and investigate reward processing mechanisms in healthy subjects \cite{perry2015reward}. However, neurodegenerative and psychiatric disorders, often associated with reward processing disruptions, can provide an additional resource for deeper understanding of human decision making mechanisms. From the perspective of evolutionary psychiatry, various mental disorders, including depression, anxiety, ADHD, addiction and even schizophrenia can be considered as ``extreme points'' in a continuous spectrum of behaviors and traits developed for various purposes during evolution, and  somewhat less extreme versions of those traits can be actually beneficial in specific environments. 
Thus, modeling decision-making biases and traits associated with various disorders may enrich the existing computational decision-making models, leading to potentially more flexible and better-performing algorithms.
In this paper, we extended previous pursuits of human behavioral agents in MAB \cite{bouneffouf2017bandit} and RL \cite{lin2019split,lin2020astory,lin2020online} into CB, built upon the Contextual Thompson Sampling (CTS) \cite{AgrawalG13}, a state-of-art approach to CB problem, and unfied all three levels as a parametric family of models, where the reward information is split into two streams,  positive and negative. 

%% file: sec_problem.tex
\section{Problem Setting}
\label{sec:problem}



\textbf{Multi-Armed Bandit (MAB).}
The multi-armed bandit (MAB) problem models a sequential decision-making process, where at each time point a player selects an action from a given finite set of possible actions, attempting to maximize the cumulative reward over time. 
Optimal solutions have been provided using a stochastic formulation \cite{LR85,UCB}, or using an adversarial formulation \cite{AuerC98,AuerCFS02,BouneffoufF16}.
Recently, there has been a surge of interest in a Bayesian formulation \cite {chapelle2011empirical}, involving the algorithm known as Thompson sampling \cite {T33}. Theoretical analysis in \cite{AgrawalG12} shows that Thompson sampling for Bernoulli bandits is asymptotically optimal. \\


\textbf{Contextual Bandit (CB).}
Following \cite{langford2008epoch}, this problem is defined as follows. At each time point (iteration) $t \in \{1,...,T\}$, an agent is presented with a {\em context} ({\em feature vector}) $\textbf{x}_t \in \mathbf{R}^N$
  before choosing an arm $k  \in A = \{ 1,...,K\} $.
We will denote by
  $X=\{X_1,...,X_N\}$  the set of features (variables) defining the context.
Let ${\textbf r_t} = (r^{1}_t,...,$ $r^{K}_t)$ denote  a reward vector, where $r^k_t \in [0,1]$ is a reward at time $t$  associated with the arm $k\in A$.
Herein, we will primarily focus on the Bernoulli bandit with binary reward, i.e. $r^k_t \in \{0,1\}$.
Let $\pi: X \rightarrow A$ denote a policy.  Also, $D_{c,r}$ denotes a joint distribution over  $({\bf x},{\bf r})$.
We will assume that the expected reward is a linear function of the context, i.e.
$E[r^k_t|\textbf{x}_t] $ $= \mu_k^T \textbf{x}_t$,
where $\mu_k$ is an unknown weight vector associated with the arm $k$. \\

\textbf{Reinforcement Learning (RL).}
Reinforcement learning defines a class of algorithms for solving problems modeled as Markov decision processes (MDP) \cite{Sutton1998}. An MDP is defined by the tuple $(\mathcal{S}, \mathcal{A}, \mathcal{T}, \mathcal{R}, \gamma)$, where  $\mathcal{S}$ is a set of possible states, $\mathcal{A}$ is a set of actions, $\mathcal{T}$ is a transition function defined as $\mathcal{T}(s, a, s')=\Pr(s'\vert s,a)$, where $s, s'\in \mathcal{S}$ and $a\in \mathcal{A}$, and $\mathcal{R}: \mathcal{S}\times \mathcal{A} \times \mathcal{S}\mapsto \mathbb{R}$ is a reward function, $\gamma$ is a discount factor that decreases the impact of the past reward on current action choice. Typically,  the objective is to maximize the discounted long-term reward, assuming  an infinite-horizon decision process, i.e. to find a policy function $\pi: \mathcal{S} \mapsto \mathcal{A}$ which specifies the action to take in a  given state, so that the cumulative reward is maximized: $\max_{\pi} \sum_{t=0}^{\infty}\gamma^t \mathcal{R}(s_t,a_t, s_{t+1}).
$

%% file: sec_method.tex
\section{Background: Contextual Thompson Sampling (CTS)}


As pointed out in the introduction, the main methodological contribution of this work is two-fold: (1) fill in the missing piece of split reward processing in the contextual bandit problem, and (2) unify the bandits, contextual bandits, and reinforcement learnings under the same framework of split reward processing mechanism. We first introduce the theoretical model we built upon for the contextual bandit problem: the Contextual Thompson Sampling, due to its known empirical benefits. In the general Thompson Sampling, the reward $r^{i}_t$ for choosing action $i$ at time $t$ follows a parametric likelihood function $Pr(r_t|\tilde{\mu}_i)$. Following \cite{AgrawalG13}, the posterior distribution at time $t + 1$, $Pr(\tilde{\mu}_i|r_t) \propto Pr(r_t|\tilde{\mu}_i) Pr(\tilde{\mu}_i)$ is given by a multivariate Gaussian distribution $\mathcal{N}(\hat{\mu}_i(t+1)$, $v^2 B_i(t + 1)^{-1})$, where
$B_i(t)= I_d + \sum^{t-1}_{\tau=1} x_{\tau} x_{\tau}^\top$, and where $d$ is the context size ${\bf x}_i$, $v= R \sqrt{\frac{24}{\epsilon} d ln(\frac{1}{\gamma})}$ with $R>0$,  $\epsilon \in ]0,1]$, $\gamma \in ]0,1]$ constants, and $\hat{\mu}_i(t)=B_i(t)^{-1} (\sum^{t-1}_{\tau=1} x_{\tau} r_{\tau})$. At every step $t$, the algorithm generates a $d$-dimensional sample $\tilde{\mu_i}$ from
$\mathcal{N}(\hat{\mu_i}(t)$, $ v^2B_i(t)^{-1})$, selects the arm $i$ that maximizes $x_t^\top \tilde{\mu_i}$, and obtains reward $r_t$.

\section{Two-Stream Split Models in MAB, CB and RL }
\label{sec:method}

We now outlined the split models evaluated in our three settings: the MAB case with the Human-Based Thompson Sampling (HBTS) \cite{bouneffouf2017bandit}, the CB case with the Split Contextual Thompson Sampling (SCTS), and the RL case with the Split Q-Learning \cite{lin2019split,lin2020astory}. All three split agent classes are standardized for their parametric notions (see Table \ref{tab:parameter} for a complete parametrization and Appendix \ref{sec:neuro} for more literature review of these clinically-inspired reward-processing biases).
\\

\textbf{Split Multi-Armed Bandit Model.}
The split MAB agent is built upon Human-Based Thompson Sampling (HBTS, algorithm \ref{alg:HBTS}) \cite{bouneffouf2017bandit}. The positive and negative streams are each stored in the success and failure counts $S_a$ and $F_a$. 

\begin{algorithm}[h!]
 \caption{\textbf{Split MAB:} Human-Based Thompson Sampling (HBTS) }
\label{alg:HBTS}
\begin{algorithmic}[1]
 \STATE {\bfseries }\textbf{Initialize:} $S_{a'} = 1$, $F_{a'} = 1, \forall a' \in A$.
  \STATE \textbf{For} each episode $e$ \textbf{do}
 \STATE {\bfseries } \quad Initialize state $s$
 \STATE {\bfseries } \quad \textbf{Repeat} for each step $t$ of the episode $e$
  \STATE {\bfseries }  \quad \quad Sample $\theta_{a'} \sim Beta(S_{a'}, F_{a'}), \forall a' \in A_t$ 
 \STATE {\bfseries } \quad \quad Take action $a= \arg \max_{a'} \theta_{a'}$, and
  \STATE {\bfseries } \quad \quad Observe $r^+ \text{ and } r^- \in R_{a'}$ \\
 \STATE {\bfseries } \quad \quad $S_{a}:=\lambda_+ S_{a}+ w_{+} r^+ $ 
 \STATE {\bfseries } \quad \quad $F_{a}:=\lambda_- F_{a}- w_{-} r^- $
\STATE {\bfseries } \quad \textbf{until} s is the terminal state
 \STATE {\bfseries }\textbf{End for}
 \end{algorithmic}
\end{algorithm}
\vspace{-0.5em}


\textbf{Split Contextual Bandit Model.}
Similarly, we now extend Contextual Thompson Sampling (CTS) \cite{AgrawalG13} to a more flexible framework, inspired by a wide range of reward-processing biases discussed in Appendix \ref{sec:neuro}. The proposed {\em Split CTS} (Algorithm \ref{alg:SCTS}) treats positive and negative rewards in two separate streams. It introduces four hyper-parameters which represent, for both positive and negative  streams, the reward processing weights (biases), as well as discount factors for the past rewards:  $\lambda_+ $ and $\lambda_-$ are the discount factors applied to the previously accumulated positive and negative rewards, respectively, while $w_+$ and $w_-$ represent the weights on the positive and negative rewards at the current iteration. We assume that at each step, an agent receives both positive and negative rewards, denote $r^+$ and $r^-$, respectively (either one of them can be zero, of course). As in HBTS, the two streams are independently updated.

\begin{algorithm}[th!]
   \caption{\textbf{Split CB:} Split Contextual Thompson Sampling (SCTS)}
\label{alg:SCTS}
\begin{algorithmic}[1]
 \STATE {\bfseries }\textbf{Initialize:} $B_{a'}^+= B_{a'}^-= I_d$, $ \hat{\mu_{a'}^+}=\hat{\mu_{a'}^-}= 0_d, f_{a'}^- = f_{a'}^- = 0_d, \forall a' \in A$.
  \STATE \textbf{For} each episode $e$ \textbf{do}
 \STATE {\bfseries } \quad Initialize state $s$
 \STATE {\bfseries } \quad \textbf{Repeat} for each step $t$ of the episode $e$
 \STATE \quad \quad Receive context $x_t$
  \STATE {\bfseries }  \quad \quad  Sample $\tilde{\mu}^+_{a'} \sim N(\hat{\mu^+_{a'}}, v^2 {B^+_{a'}}^{-1})$ and $\tilde{\mu}^-_{a'} \sim N(\hat{\mu^-_{a'}}, v^2 {B^-_{a'}}^{-1}), \forall a' \in A_t$ 
 \STATE {\bfseries } \quad \quad Take action $a= \arg \max_{a'} (x_t^\top \tilde{\mu}^+_{a'} + x_t^\top \tilde{\mu}^-_{a'})$, and
  \STATE {\bfseries } \quad \quad Observe $r^+ \text{ and } r^- \in R_{a'}$ \\
 \STATE {\bfseries }  \quad \quad $B^+_{a}:= \lambda_+ B^+_{a} + x_t x_t^\top $, $f^+_{a} := \lambda_+ f^+_{a} + w_{+} x_t r^+$, $\hat{\mu^+_{a}} := {B^+_{a}}^{-1} f^+_{a}$
 \STATE {\bfseries }  \quad \quad $B^-_{a}:= \lambda_- B^-_{a} + x_t x_t^\top $, $f^-_{a} := \lambda_- f^-_{a} + w_{-} x_t r^-$, $\hat{\mu^-_{a}} := {B^-_{a}}^{-1} f^-_{a}$
 \STATE {\bfseries } \quad \textbf{until} s is the terminal state
 \STATE {\bfseries }\textbf{End for}
   \end{algorithmic}
\end{algorithm}
\vspace{-0.5em}


\textbf{Split Reinforcement Learning Model.}
The split RL agent is built upon Split Q-Learning (SQL, Algorithm \ref{alg:SQL}) by \cite{lin2019split,lin2020astory} (and its variant, MaxPain, by \cite{elfwing2017parallel}). The processing of the positive and negative streams is handled by the two independently updated Q functions, $Q^+$ and $Q^-$.

\begin{algorithm}[ht!]
 \caption{\textbf{Split RL:} Split Q-Learning (SQL)}
\label{alg:SQL}
\begin{algorithmic}[1]
  \STATE {\bfseries } {\bf Initialize:} $Q$, $Q^+$, $Q^-$ tables (e.g., to all zeros)
  \STATE \textbf{For} each episode $e$ \textbf{do}
 \STATE {\bfseries } \quad Initialize state $s$
 \STATE {\bfseries } \quad \textbf{Repeat} for each step $t$ of the episode $e$
 \STATE {\bfseries }  \quad \quad  $Q(s,a') := Q^{+}(s,a') +  Q^{-}(s,a'), \forall a' \in A_t$
 \STATE {\bfseries } \quad \quad Take action $a = \arg \max_{a'}Q(s,a')$, and
  \STATE {\bfseries } \quad \quad Observe $s'\in S$, $r^+ \text{ and } r^- \in R(s), s \leftarrow s'$ \\
 \STATE {\bfseries }  \quad \quad $Q^{+}(s,a):=\lambda_+\hat{Q}^{+}(s,a)+ $ 
 $\alpha_t(w_{+}r^{+}+\gamma \max_{a'}\hat{Q}^{+}(s',a')-\hat{Q}^{+}(s,a))$
 \STATE {\bfseries } \quad \quad $Q^{-}(s,a):= \lambda_-\hat{Q}^{-}(s,a)+$ 
$ \alpha_t(w_{-}r^{-}+\gamma   \max_{a'}\hat{Q}^{-}(s',a')-\hat{Q}^{-}(s,a))$
 \STATE {\bfseries } \quad \textbf{until} s is the terminal state
 \STATE {\bfseries }\textbf{End for}
 \end{algorithmic}
\end{algorithm}

\textbf{Clinically inspired Reward Processing Biases in Split Models.}
For each agent, we set the four parameters: $\lambda_+$ and $\lambda_-$ as the weights of the previously accumulated positive and negative rewards, respectively, $w_+$ and $w_-$ as the weights on the positive and negative rewards at the current iteration. {\em DISCLAIMER: while we use  disorder names for the models, we are not claiming that the models accurately capture all aspects of the corresponding disorders.}

In the following section we describe how specific constraints on the model parameters in the proposed method can generate a range of reward processing biases, and introduce several instances of the split models associated with those biases; the corresponding parameter settings are presented in Table \ref{tab:parameter}. As we demonstrate later, specific biases may be actually beneficial in some settings, and our parameteric approach often outperforms the standard baselines due to increased generality and flexibility of our two-stream, multi-parametric formulation.

Note that the {\em standard} split approach correspond to setting the four (hyper)parameters used in our model to 1. We also introduce two variants which only learn from one  of the two reward streams: negative split models (algorithms that start with N) and positive split models (algorithms that start with P), by setting to zero $ \lambda_+$ and $w_+$, or $ \lambda_-$ and $w_-$, respectively. Next, we introduce the model which incorporates some mild forgetting of the past rewards or losses (0.5 weights) and calibrating the other models with respect to this one; we refer to this model as M for ``moderate'' forgetting. 

We also specified the mental agents differently with the prefix ``b-'' referring to the MAB version of the split models (as in ``bandits'), ``cb-'' referring to the CB version, and no prefix as the RL version (for its general purposes).

We will now introduced several models inspired by certain reward-processing biases in a range of mental disorders-like behaviors in table \ref{tab:parameter}.

Recall that PD patients are typically better at learning to avoid negative outcomes than at learning to achieve positive outcomes \cite{frank2004carrot}; one way to model this is to over-emphasize negative rewards, by placing a high weight on them, as compared to the reward processing in healthy individuals. Specifically, we will assume the parameter $w_-$ for PD patients to be much higher than normal $w_-$ (e.g., we use $w_-=100$ here), while the rest of the parameters will be in the same range for both healthy and PD individuals.
Patients with bvFTD are prone to overeating which may represent increased reward representation. To model this impairment in bvFTD patients, the parameter of the model could be modified as follow: $w_+^M <<w_+$ (e.g.,  $w_+=100$ as shown in Table \ref{tab:parameter}), where $w_+$ is the parameter of the bvFTD model has, and the rest of these parameters are equal to the normal one.
To model apathy in patients with Alzheimer's, including downplaying rewards and losses, we will assume that the parameters $\lambda_+$ and $\lambda_-$ are  somewhat smaller than normal, $ \lambda_+ < \lambda_+^M$ and $\lambda_- < \lambda_-^M $ (e.g,  set to 0.1 in Table \ref{tab:parameter}), which models the tendency to  forget both positive and negative rewards.
 Recall that ADHD may be involve impairments in storing stimulus-response associations. In our ADHD model, the parameters $\lambda_+$ and $\lambda_-$ are smaller than normal, $\lambda_+^M > \lambda_+$ and $\lambda_-^M > \lambda_-$, which models forgetting of both positive and negative rewards. Note that while this model appears similar to Alzheimer's model described above, the forgetting factor will be  less pronounced, i.e. the $\lambda_+$ and $\lambda_-$ parameters are larger than those of the  Alzheimer's model (e.g., 0.2 instead of 0.1, as shown in Table  \ref{tab:parameter}).
 As mentioned earlier, addiction is associated with inability to properly forget (positive) stimulus-response associations; we model this by setting the weight on previously accumulated positive reward (``memory'' )  higher than normal, $\tau >\lambda_+^M $, e.g. $\lambda_+ = 1$, while $\lambda_+^M = 0.5$. We model the reduced responsiveness to rewards in chronic pain by setting $ w_+ < w_+^M $ so there is a decrease in the reward representation, and $\lambda_- > \lambda_-^M $ so the negative rewards are not forgotten (see table \ref{tab:parameter}).

Of course, the above models should be treated only as first approximations of the reward processing  biases in mental disorders, since the actual changes in reward processing are much more complicated, and the parameteric setting must be learned from actual patient data, which is a nontrivial direction for future work. Herein, we simply consider those models as specific variations of our general method, inspired by certain aspects of the corresponding diseases, and focus primarily on the computational aspects of our algorithm, demonstrating that the proposed parametric extension of standard algorithms can learn better than the baselines due to added flexibility.

\begin{table}[tb]
\centering
\caption{\textbf{Parameter setting} for different types of reward biases in the split models.}
\label{tab:parameter}
\resizebox{0.8\columnwidth}{!}{
 \begin{tabular}{ l | c | c | c | c }
  & $\lambda_+$   & $w_+$     & $\lambda_-$      & $w_-$ \\ \hline
  ``Addiction'' (ADD)   & $1 \pm 0.1$   & $1 \pm 0.1$    & $0.5 \pm 0.1$   & $1 \pm 0.1$ \\
  ``ADHD''  & $0.2\pm 0.1$  & $1 \pm 0.1$    & $0.2 \pm 0.1$   & $1 \pm 0.1$ \\
  ``Alzheimer's'' (AD)   & $0.1 \pm 0.1$  & $1 \pm 0.1$    & $0.1 \pm 0.1$   & $1 \pm 0.1$ \\
  ``Chronic pain'' (CP)   & $0.5 \pm 0.1$  & $0.5 \pm 0.1$    & $1 \pm 0.1$    & $1 \pm 0.1$ \\
  ``bvFTD''  & $0.5 \pm 0.1$  & $100 \pm 10$    & $0.5 \pm 0.1$   & $1 \pm 0.1$ \\
  ``Parkinson's'' (PD)   & $0.5 \pm 0.1$  & $1 \pm 0.1$    & $0.5\pm 0.1$   & $100\pm 10$\\
  ``moderate'' (M)   & $0.5 \pm 0.1$  & $1 \pm 0.1$    & $0.5 \pm 0.1$   & $1 \pm 0.1$ \\
  \hline 
  Standard (HBTS, SCTS, SQL)  & 1        & 1         & 1         & 1\\
  Positive (PTS, PCTS, PQL) & 1        & 1         & 0         & 0\\
  Negative (NTS, NCTS, NQL)  & 0        & 0         & 1         & 1\\
 \end{tabular}
 }
    \vspace{0.1in}
 \end{table}

\clearpage

%% file: sec_experiment.tex
\vspace{-1em}
\section{Empirical Evaluation}
\label{sec:results}

\input{./sec_tables.tex}

Empirically, we evaluated the algorithms in four settings: the gambling game of a simple MDP task, a simple MAB task, a real-life Iowa Gambling Task (IGT) \cite{steingroever2015data}, and a PacMan game. There is considerable randomness in the reward, and predefined multimodality in the reward distributions of each state-action pairs in all four tasks. We ran split MAB agents in MAB, MDP and IGT tasks, and split CB and RL agents in all four tasks. \\



\subsection{MAB and MDP Tasks with bimodal rewards} 

In this simple MAB example, a player starts from initial state A, choose between two actions: go left to reach state B, or go right to reach state C. Both states B and C reveals a zero rewards. From state B, the player observes a reward from a distribution $R_B$. From state C, the player observes a reward from a distribution $R_C$. The reward distributions of states B and C are both multimodal distributions (for instance, the reward $r$ can be drawn from a bi-modal distribution of two normal distributions $N(\mu=10,\sigma=5)$ with probability $p=0.3$ and $N(\mu=-5,\sigma=1)$ with $p=0.7$). The left action (go to state B) by default is set to have an expected payout lower than the right action. However, the reward distributions can be spread across both the positive and negative domains. For Split models, the reward is separated into a positive stream (if the revealed reward is positive) and a negative stream (if the revealed reward is negative).\\

\textbf{Experiments.}
To evaluate the robustness of the algorithms, we simulated 100 randomly generated scenarios of bi-modal distributions, where the reward can be drawn from two normal distribution with means as random integers uniformly drawn from -100 to 100, standard deviations as random integers uniformly drawn from 0 to 50, and sampling distribution $p$ uniformly drawn from 0 to 1 (assigning $p$ to one normal distribution and $1-p$ to the other one). Each scenario was repeated 50 times with standard errors as bounds. In all experiments, the discount factor $\gamma$ was set to be 0.95. For non-exploration approaches, the exploration is included with $\epsilon$-greedy algorithm with $\epsilon$ set to be 0.05. The learning rate was polynomial $\alpha_t(s, a) = 1/n_t(s, a)^{0.8}$, which is better in theory and in practice \cite{even2003learning}. 
\\

\textbf{Benchmark.}
We compared the following algorithms: In MAB setting, we have Thompson Sampling (TS) \cite{T33}, Upper Confidence Bound (UCB) \cite{UCB}, epsilon Greedy (eGreedy) \cite{Sutton98}, EXP3 \cite{auer2002nonstochastic} (and gEXP3 for the pure greedy version of EXP3), Human Based Thompson Sampling (HBTS) \cite{bouneffouf2017bandit}. In CB setting, we have Contextual Thompson Sampling (CTS) \cite{AgrawalG13}, LinUCB \cite{LiCLW11}, EXP4 \cite{beygelzimer2011contextual} and Split Contextual Thompson Sampling (SCTS). In RL setting, we have Q-Learning (QL), Double Q-Learning (DQL) \cite{hasselt2010double}, State–action–reward–state–action (SARSA) \cite{rummery1994line}, Standard Split Q-Learning (SQL) \cite{lin2019split,lin2020astory}, MaxPain (MP) \cite{elfwing2017parallel}, Positive Q-Learning (PQL) and Negative Q-Learning (NQL). 
\\

\begin{figure*}[tb]
\vspace{-1em}
\centering
    \includegraphics[width=\linewidth]{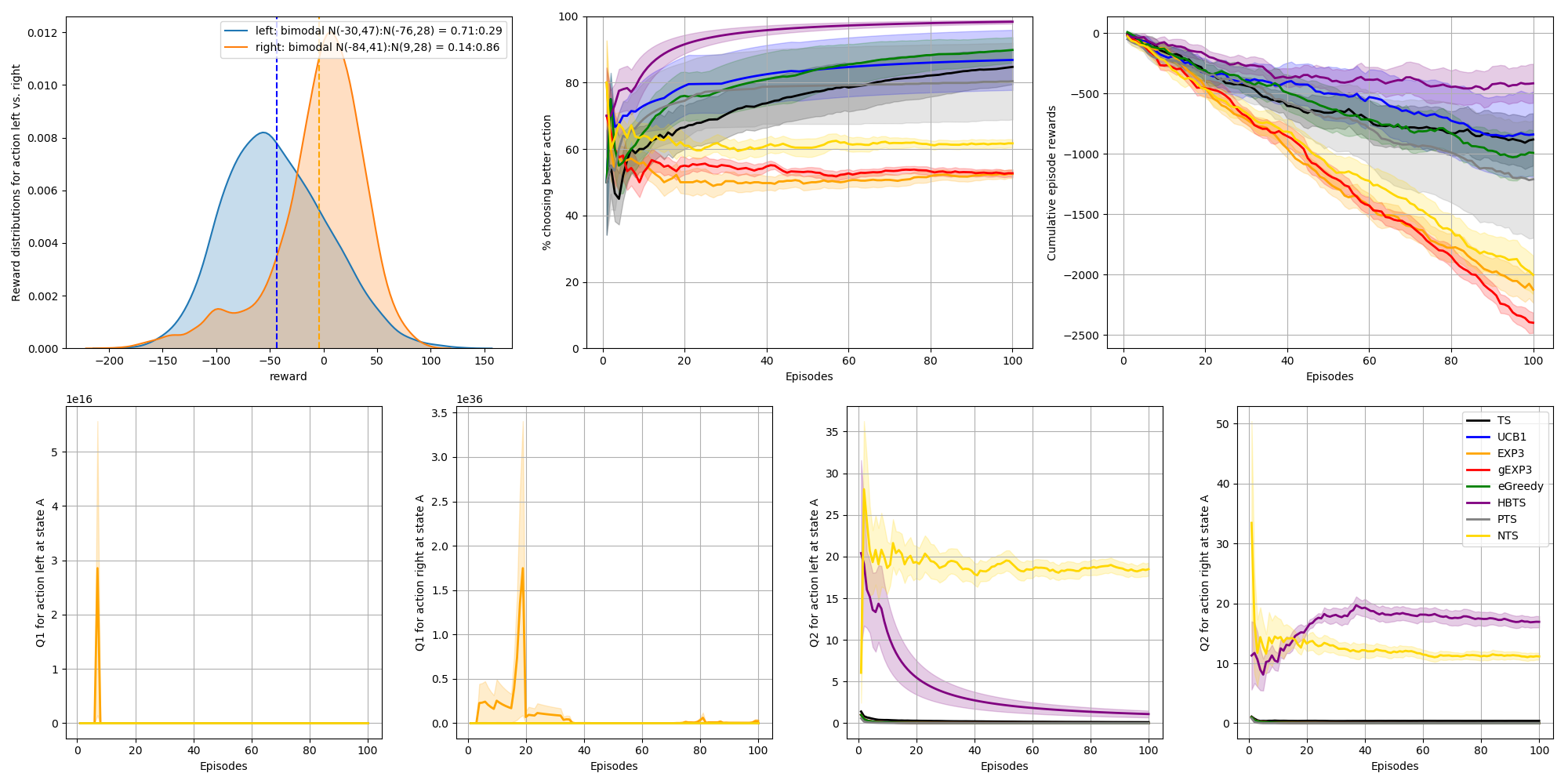}
        \vspace{-1em}
\par\caption{\textbf{MAB in MAB task:} example where Split MAB performs better than baselines.}\label{fig:MAB2}
    \vspace{-0.5em}
\end{figure*}

\begin{figure*}[tb]
\vspace{-1em}
\centering
    \includegraphics[width=\linewidth]{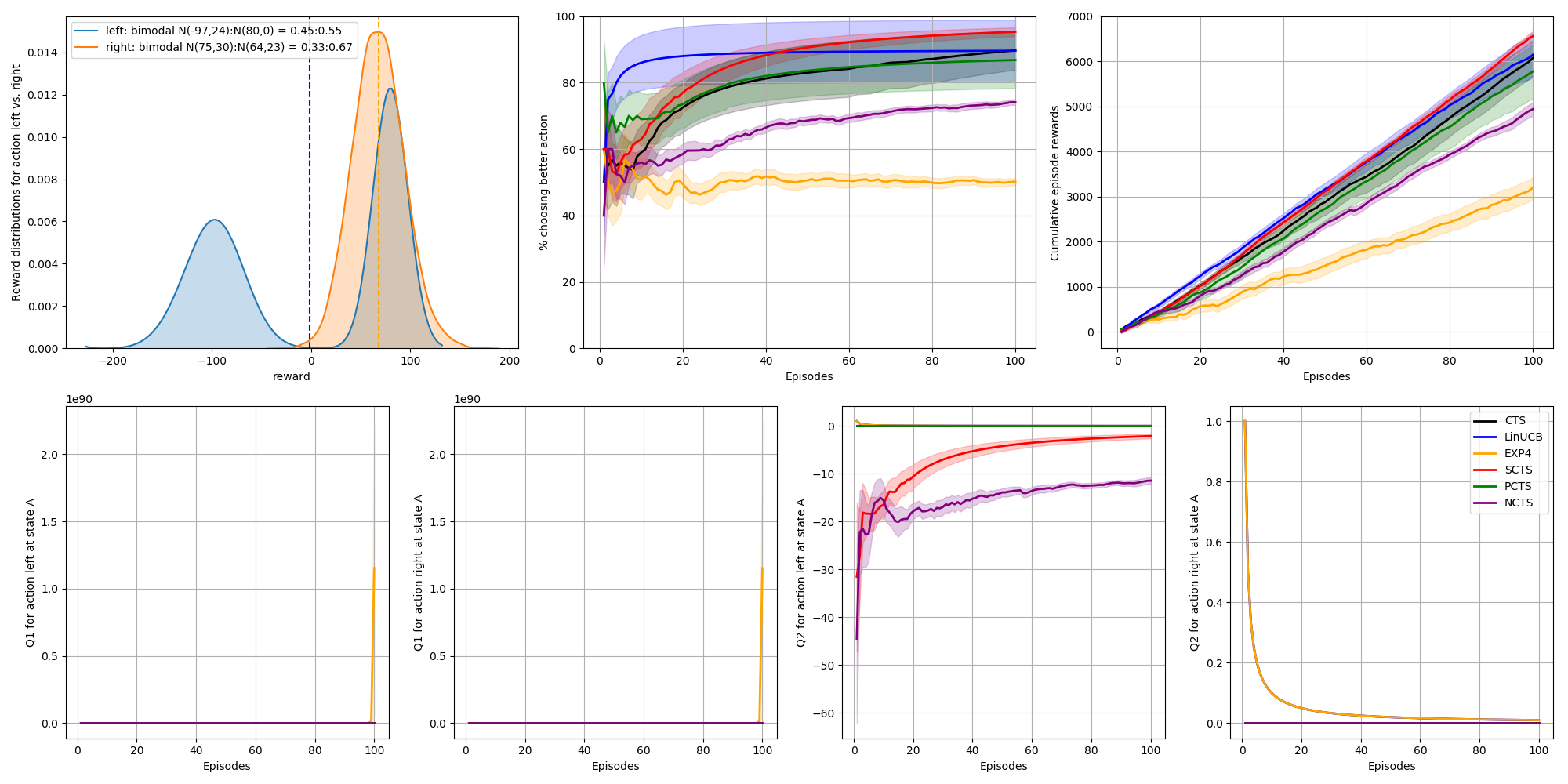}
        \vspace{-1em}
\par\caption{\textbf{CB in MAB task:} example where SCTS performs better than baselines.}\label{fig:MAB}
    \vspace{-0.5em}
\end{figure*}

\textbf{Evaluation Metric.}
In order to evaluate the performances of the algorithms, we need a scenario-independent measure which is not dependent on the specific selections of reward distribution parameters and pool of algorithms being considered. The final cumulative rewards might be subject to outliers because they are scenario-specific. The ranking of each algorithms might be subject to selection bias due to different pools of algorithms being considered. The pairwise comparison of the algorithms, however, is independent of the selection of scenario parameters and selection of algorithms. For example, in the 100 randomly generated scenarios, algorithm X beats Y for $n$ times while Y beats X $m$ times. We may compare the robustness of each pairs of algorithms with the ratio $n:m$.
 
\textbf{Results.}
Figure \ref{fig:MAB2} and Figure \ref{fig:MAB} are two example scenarios plotting the reward distributions, the percentage of choosing the better action (go right), the cumulative rewards and the changes of two Q-tables (the weights stored in $\tilde{\mu}_a^+$ and $\tilde{\mu}_a^-$) over the number of iterations, drawn with standard errors over multiple runs. Each trial consisted of a synchronous update of all 100 actions. With polynomial learning rates, we see split models (HBTS in bandit agent pool, SCTS in contextual bandit agent pool, and SQL in RL agent pool) converged much more quickly than baselines.

Tables 
\ref{tab:MAB} and \ref{tab:MDP}
summarized the pairwise comparisons between the agents with the row labels as the algorithm X and column labels as algorithm Y giving $n:m$ in each cell denoting X beats Y $n$ times and Y beats X $m$ times. For each cell of $i$th row and $j$th column, the first number indicates the number of rounds the agent $i$ beats agent $j$, and the second number the number of rounds the agent $j$ beats agent $i$. The average wins of each agent is computed as the mean of the win rates against other agents in the pool of agents in the rows. The bold face indicates that the performance of the agent in column $j$ is the best among the agents, or the better one. Among the algorithms, split models never seems to fail catastrophically by maintaining an overall advantages over the other algorithms. 

For instance, in the MAB task, among the MAB agent pool, HBTS beats non-split version of TS with a winning rate of 52.65\% over 46.72\%. In the CB agent pool, LinUCB performed the best with a winning rate of 57.07\%. This suggested that upper confidence bound (UCB)-based approach are more suitable for the two-armed MAB task that we proposed, although theoretical analysis in \cite{AgrawalG12} shows that Thompson sampling models for Bernoulli bandits are asymptotically optimal. Further analysis is worth pursuing to explore UCB-based split models. In the RL agent pool, we observe that SARSA algorithm is the most robust among all agents, suggesting a potential benefit of the on-policy learning in the two-armed MAB problem that we proposed. Similarly in the MDP task, the behavior varies. In the MAB agent pool, despite not built with state representation, gEXP, an adversarial bandit algorithm with the epsilon greedy exploration performed the best. We suspected that our non-Gaussian reward distribution might resemble the nonstationary or adversarial setting that EXP3 algorithm is designed for. In the CB agent pool, we observed that LinUCB performed the best, which matched our finding in the similar MAB task above. In the RL agent pool, one of the split models, MP performed the best against all baselines, suggesting a benefit in the split mechanism in the MDP environments that we generated.  

To explore the variants of split models representing different mental disorders, we also performed the same experiments on the 7 disease models proposed above. Tables 
\ref{tab:MABmental} and \ref{tab:MDPmental} 
summarized their pairwise comparisons with the standard ones, where the average wins are computed averaged against three standard baseline models. Overall, PD (``Parkinson's''), CP (``chronic pain''), ADHD and M (``moderate'') performed relatively well. In the MAB setting, the optimal reward bias are PD and M for the split MAB models, ADHD and CP for the split CB models, and bvFTD and M for the split RL models. In the MDP setting, the optimal reward bias are PD and M for the split MAB models, ADHD and bvFTD for the split CB models, and ADHD and CP for the split RL models.






\clearpage

\begin{figure*}[tb]
\centering
    \begin{minipage}{0.08\linewidth}MAB\end{minipage}
\begin{minipage}{0.9\linewidth}
\includegraphics[width=\linewidth]{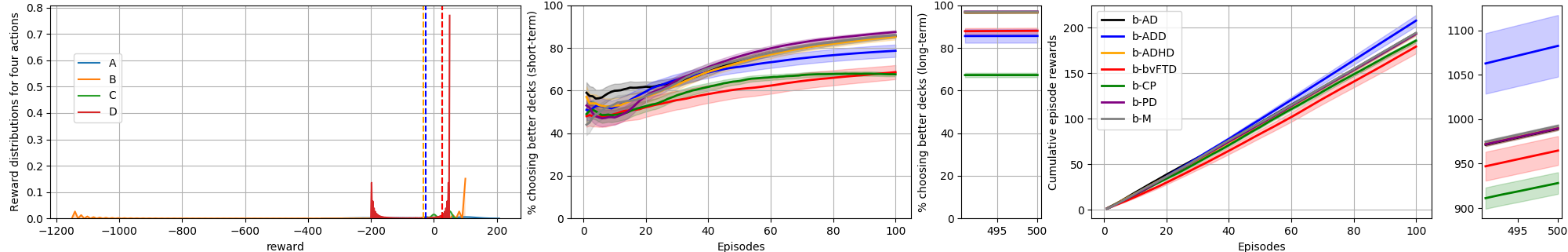}
    \end{minipage}
\begin{minipage}{0.08\linewidth}CB\end{minipage}
\begin{minipage}{0.9\linewidth}
\includegraphics[width=\linewidth]{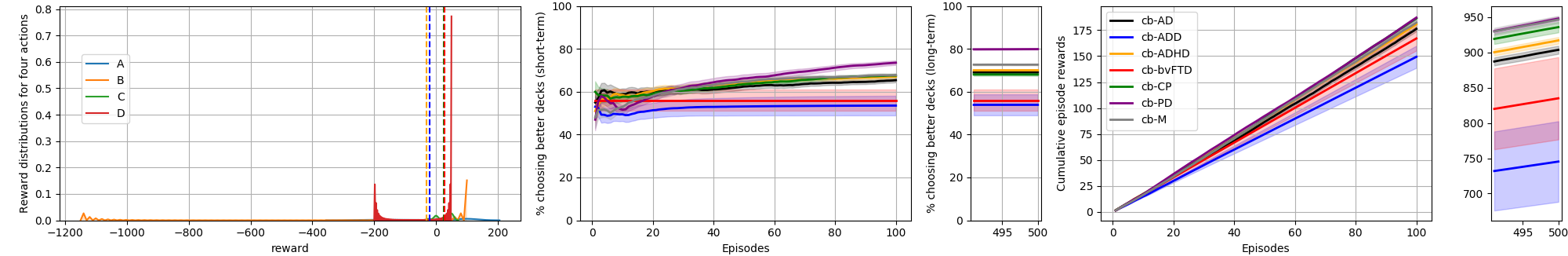}
    \end{minipage}
\begin{minipage}{0.08\linewidth}RL\end{minipage}
\begin{minipage}{0.9\linewidth}
\includegraphics[width=\linewidth]{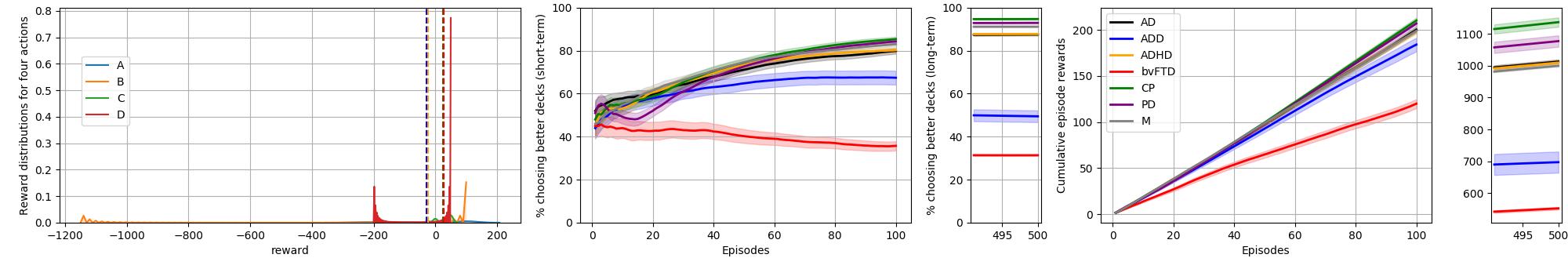}
\end{minipage}
    \par\caption{\textbf{Learning curves in IGT scheme 1:} ``Mental'' MAB, CB and RL agents.}\label{fig:IGT}
\end{figure*}

 \begin{table*}[tb]
\centering
\caption{\textbf{Schemes} of Iowa Gambling Task}
\label{tab:IGTschemes}
\resizebox{1\linewidth}{!}{
 \begin{tabular}{ l | c | l | c | c }
  Decks & win per card  & loss per card & expected value & scheme \\ \hline
  A (bad) & +100 & Frequent: -150 (p=0.1), -200 (p=0.1), -250 (p=0.1), -300 (p=0.1), -350 (p=0.1) & -25 & 1 \\
  B (bad) & +100 & Infrequent: -1250 (p=0.1) & -25 & 1 \\  
  C (good) & +50 & Frequent: -25 (p=0.1), -75 (p=0.1),-50 (p=0.3) & +25 & 1 \\  
  D (good) & +50 & Infrequent: -250 (p=0.1) & +25 & 1 \\   \hline
  A (bad) & +100 & Frequent: -150 (p=0.1), -200 (p=0.1), -250  (p=0.1), -300 (p=0.1), -350 (p=0.1) & -25 & 2 \\
  B (bad) & +100 & Infrequent: -1250 (p=0.1) & -25 & 2 \\  
  C (good) & +50 & Infrequent: -50 (p=0.5) & +25 & 2 \\  
  D (good) & +50 & Infrequent: -250 (p=0.1) & +25 & 2 \\  
 \end{tabular}
 } 
 \end{table*}
 
\subsection{Iowa Gambling Task} 

The original Iowa Gambling Task (IGT) studies decision making where the participant needs to choose one out of four card decks (named A, B, C, and D), and can win or lose money with each card when choosing a deck to draw from \cite{bechara1994insensitivity}, over around 100 actions. In each round, the participants receives feedback about the win (the money he/she wins), the loss (the money he/she loses), and the combined gain (win minus lose). In the MDP setup, from initial state I, the player select one of the four deck to go to state A, B, C, or D, and reveals positive reward $r^+$ (the win), negative reward $r^-$ (the loss) and combined reward $r=r^++r^-$ simultaneously. Decks A and B by default is set to have an expected payout (-25) lower than the better decks, C and D (+25). For baselines, the combined reward $r$ is used to update the agents. For split models, the positive and negative streams are fed and learned independently given the $r^+$ and $r^-$.

There are two major payoff schemes in IGT. In the traditional payoff scheme, the net outcome of every 10 cards from the bad decks (i.e., decks A and B) is -250, and +250 in the case of the good decks (i.e., decks C and D). There are two decks with frequent losses (decks A and C), and two decks with infrequent losses (decks B and D). All decks have consistent wins (A and B to have +100, while C and D to have +50) and variable losses (summarized in Table \ref{tab:IGTschemes}, where scheme 1 \cite{fridberg2010cognitive} has a more variable losses for deck C than scheme 2 \cite{horstmann2012iowa}). We performed the each scheme for 200 times over 500 actions. 

\begin{figure*}[tb]
\centering
\begin{minipage}{0.03\linewidth}
CB
    \end{minipage}
\begin{minipage}{0.95\linewidth}
    \includegraphics[width=0.24\linewidth]{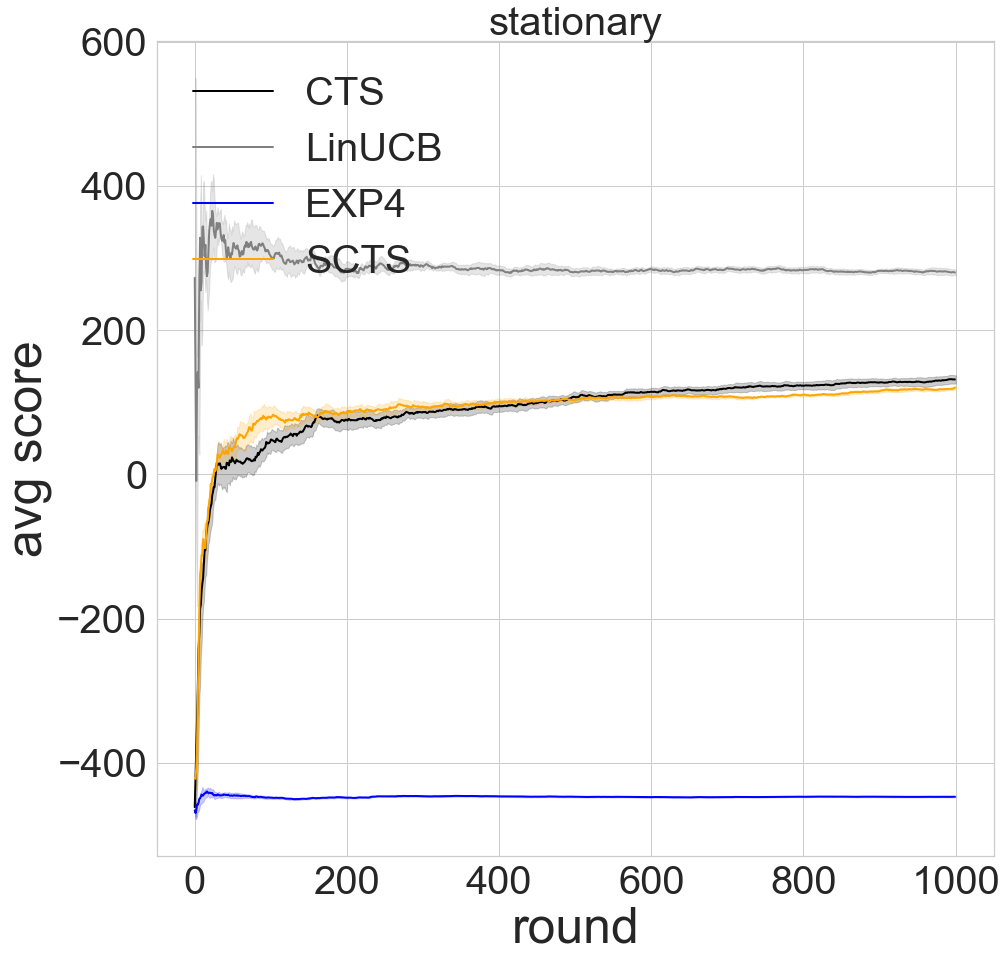}
    \includegraphics[width=0.24\linewidth]{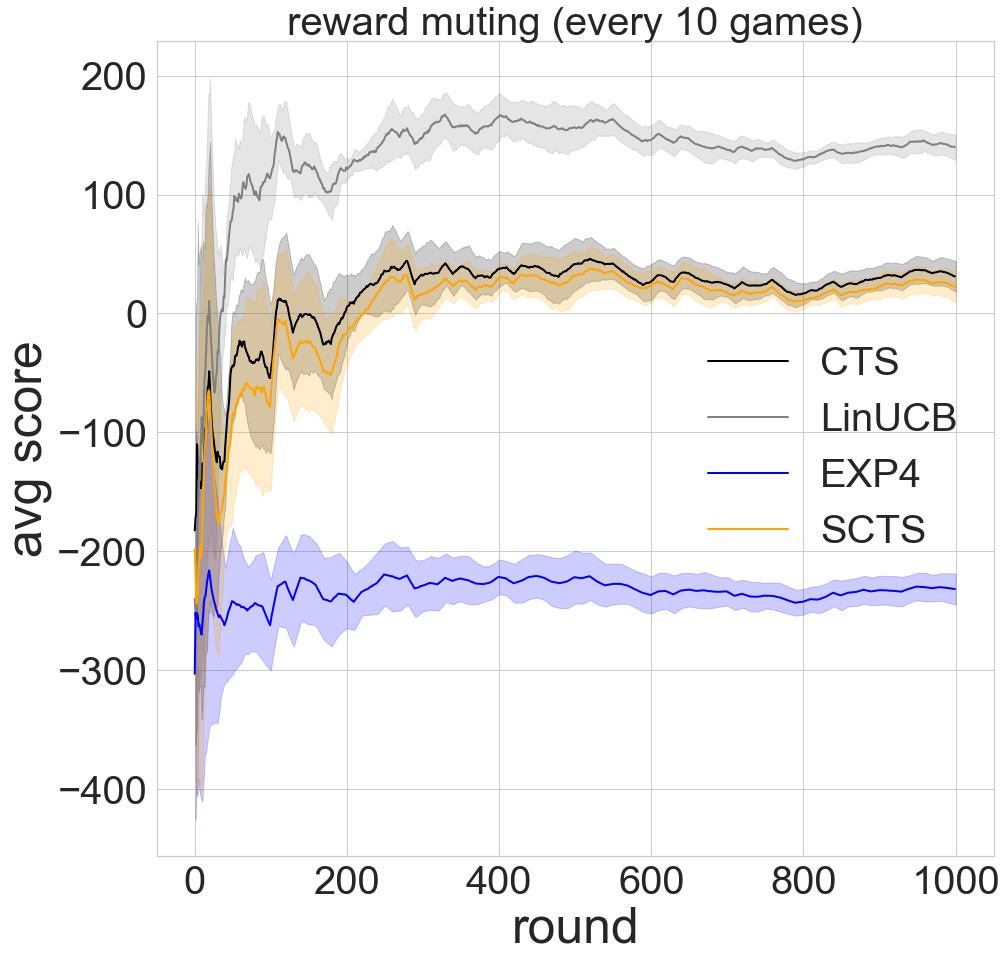}
    \includegraphics[width=0.24\linewidth]{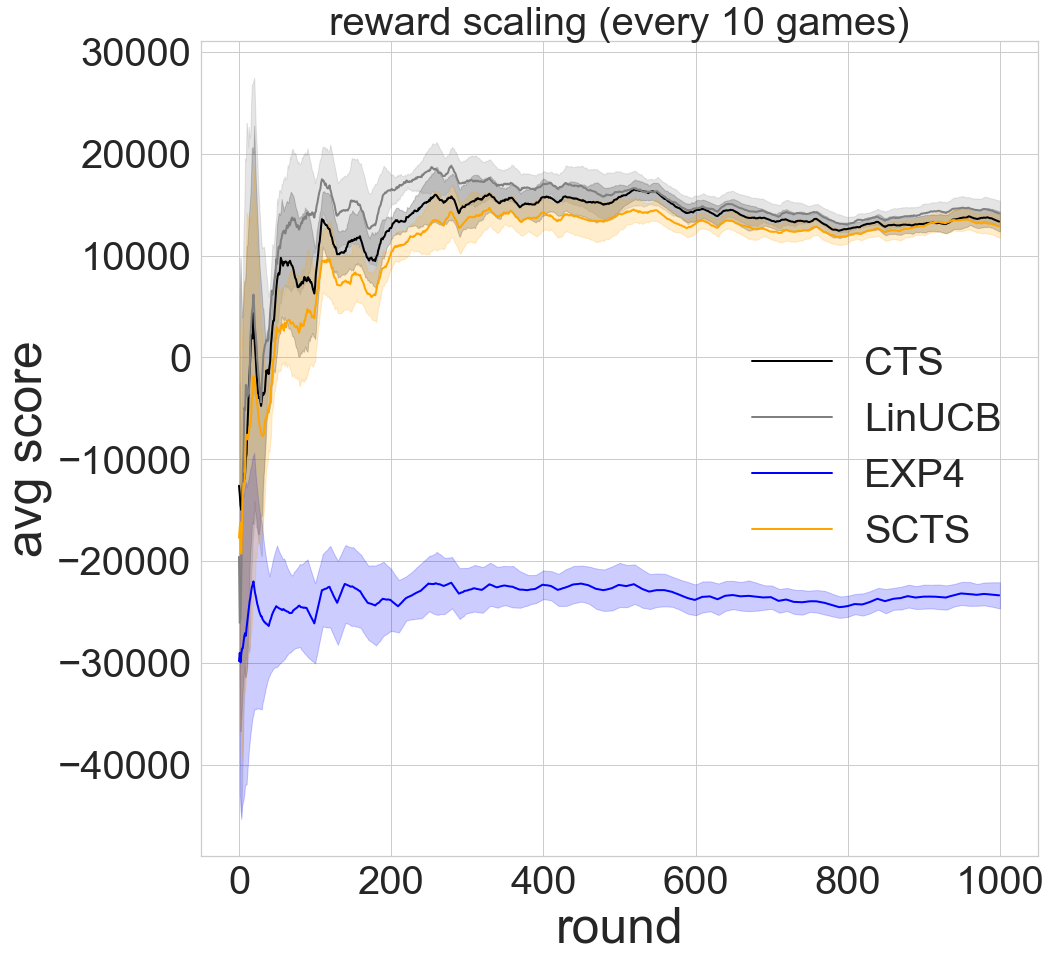}
    \includegraphics[width=0.24\linewidth]{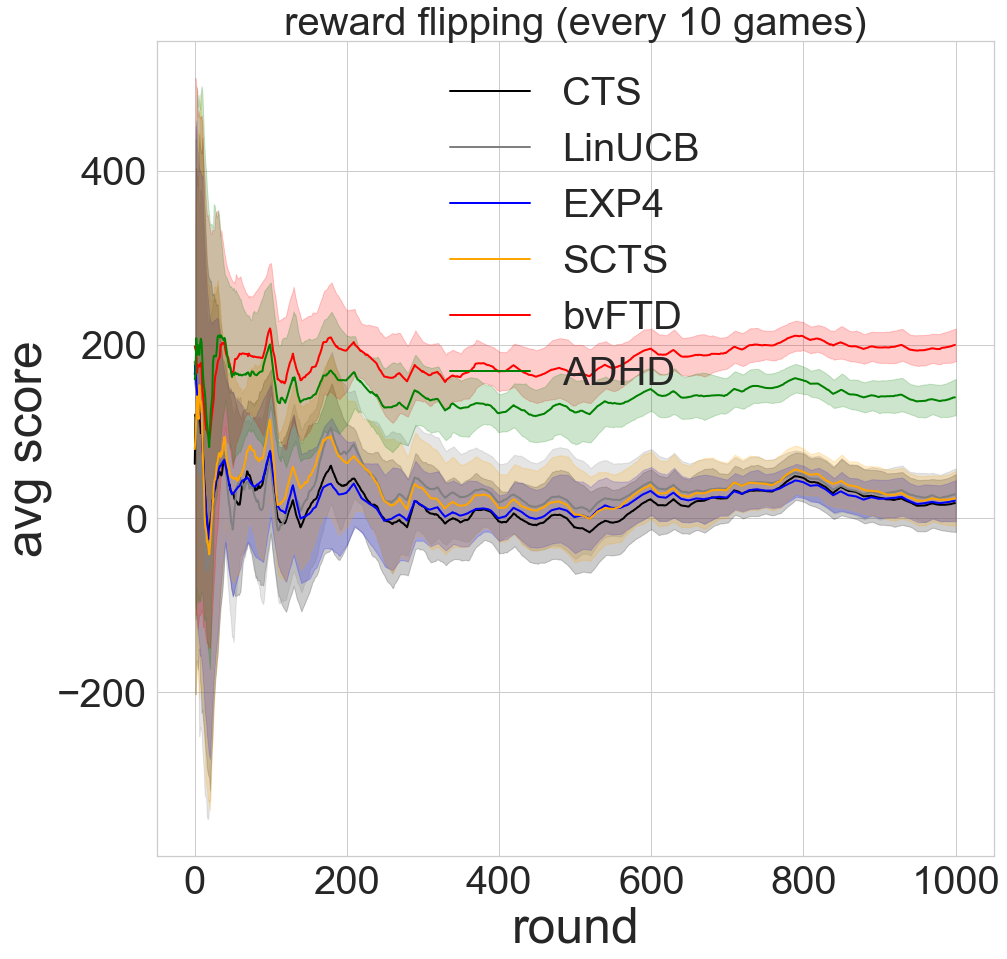}    \end{minipage}
    \begin{minipage}{0.03\linewidth}
RL
    \end{minipage}
\begin{minipage}{0.95\linewidth}
    \includegraphics[width=0.24\linewidth]{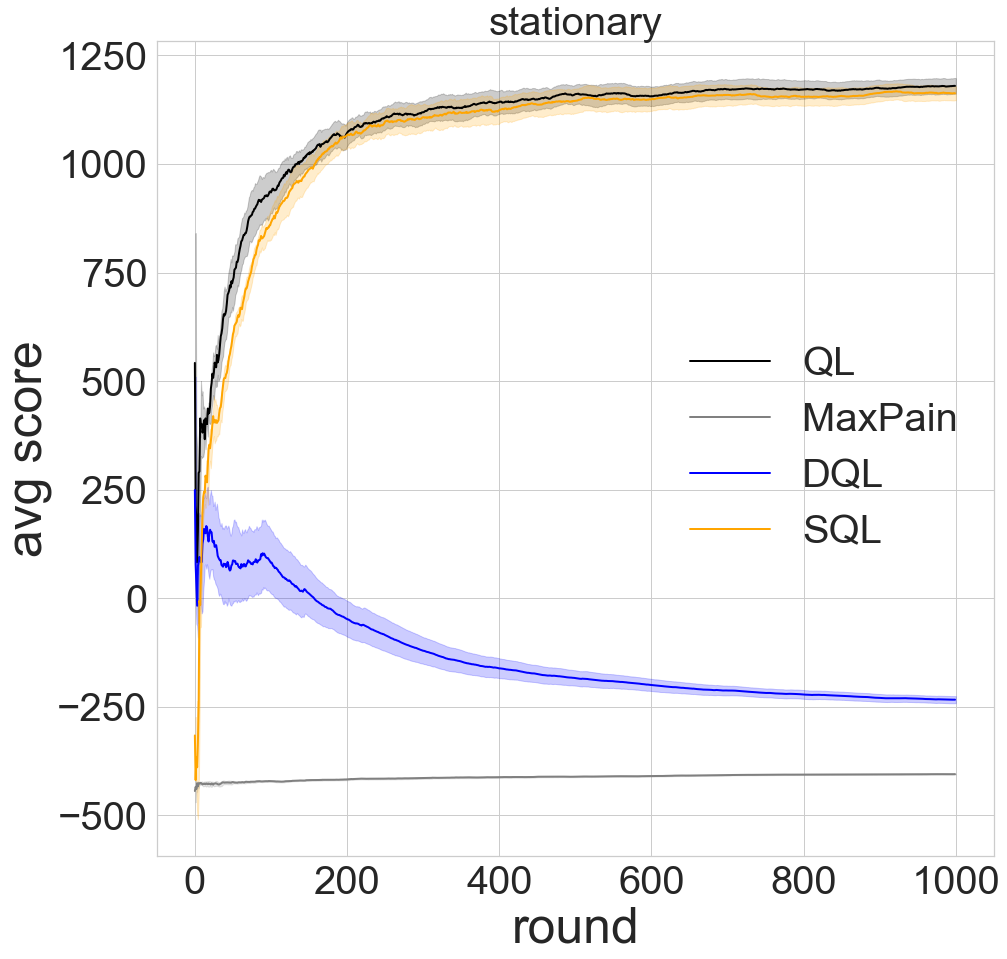}
    \includegraphics[width=0.24\linewidth]{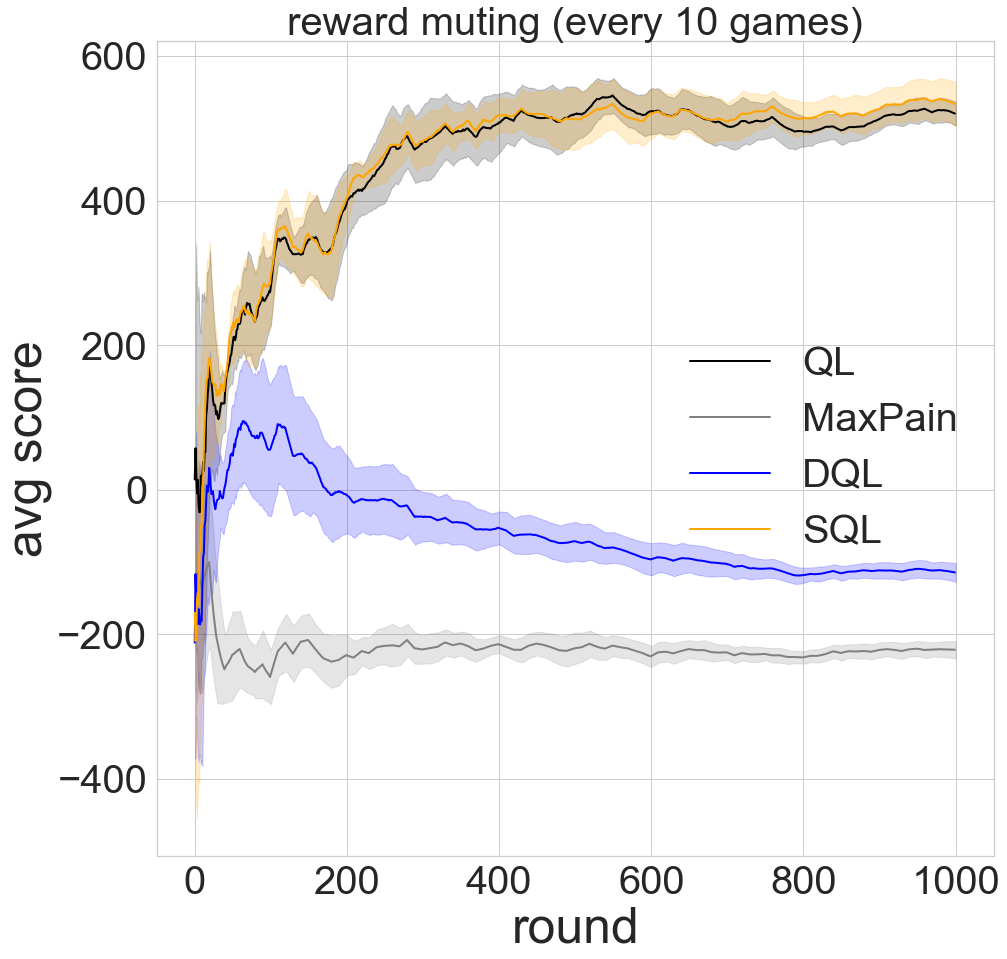}
    \includegraphics[width=0.24\linewidth]{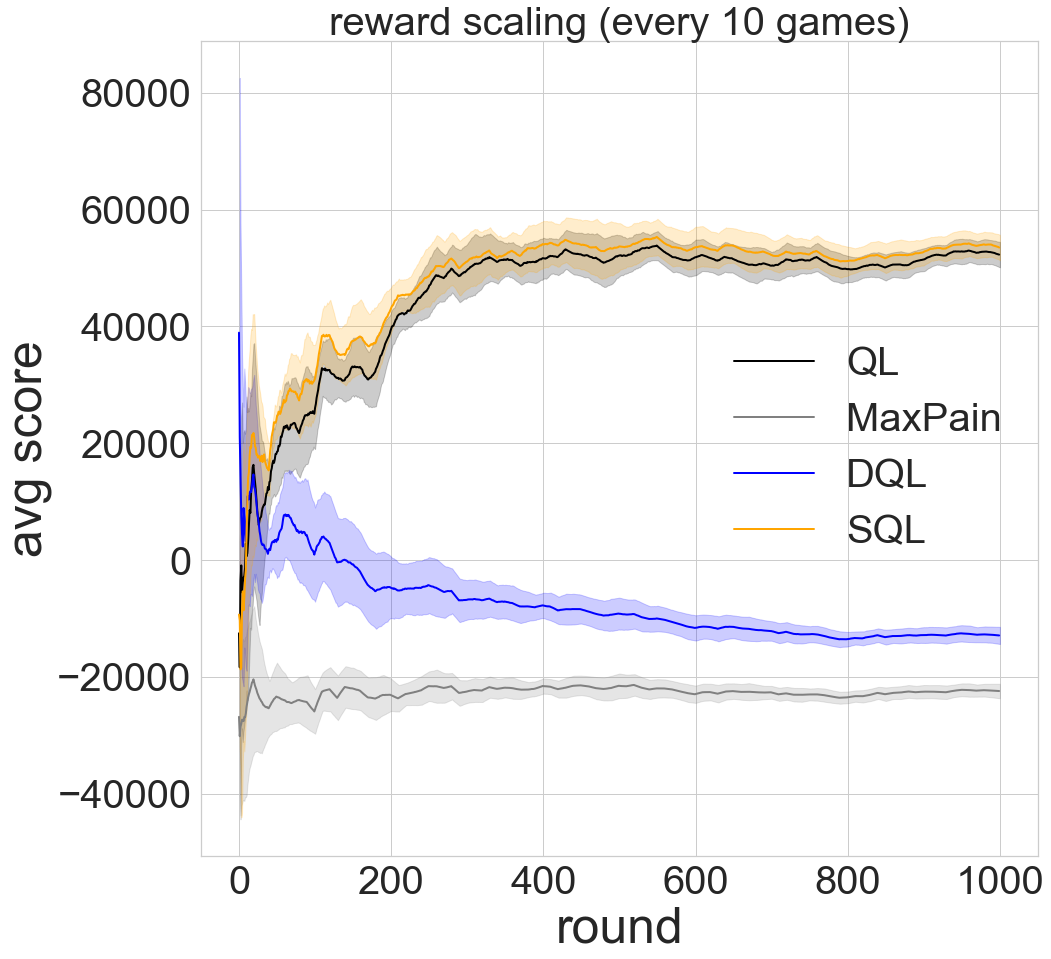}
    \includegraphics[width=0.24\linewidth]{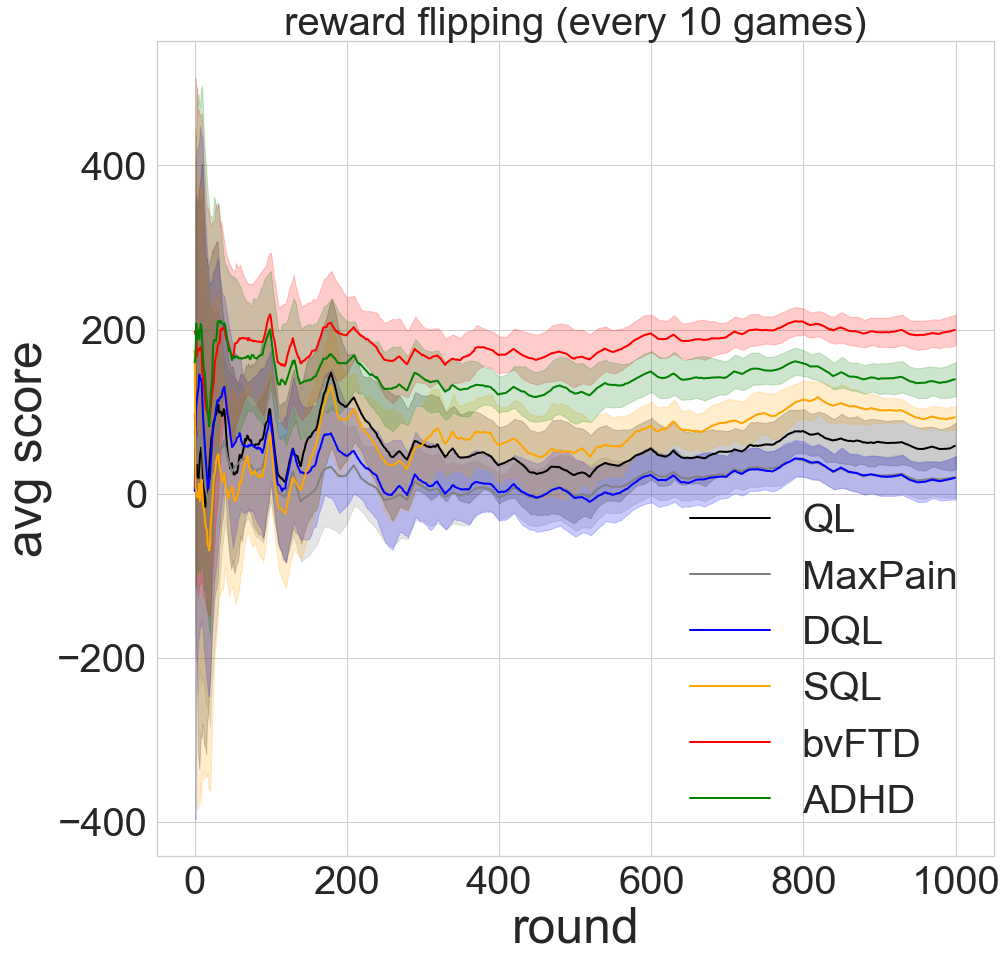}    \end{minipage}
    \par\caption{\textbf{Average final scores in Pacman with different stationarities:} Columns as (a) stationary; (b) stochastic reward muting by every 10 rounds; (c) stochastic reward scaling by every 10 rounds; (d) stochastic reward flipping by every 10 rounds.}\label{fig:pacman_results}
\end{figure*}

\textbf{Results.} 
 Among the variants of Split models and baselines, the split contextual bandit (SCTS) performs best in scheme 1 with an averaged final cumulative rewards of 1200.76 over 500 draws of cards, significantly better than the MAB baseline TS (991.26), CB baseline LinUCB (1165.23) and RL baseline QL (1086.33). Mental variants of SCTS, such as CP (``chronic pain'', 1136.38), also performed quite well. This is consistent to the clinical implication of chronic pain patients which tend to forget about positive reward information (as modeled by a smaller $\lambda_+$) and lack of drive to pursue rewards (as modeled by a smaller $w_+$). In scheme 2, eGreedy performs best with the final score of 1198.32, followed by CP (1155.84) and SCTS (1150.22). These examples suggest that the proposed framework has the flexibility to map out different behavior trajectories in real-life decision making (such as IGT). Figure \ref{fig:IGT} demonstrated the short-term (in 100 actions) and long-term behaviors of different mental agents, which matches clinical discoveries. For instance, ADD (``addiction'') quickly learns about the actual values of each decks (as reflected by the short-term curve) but in the long-term sticks with the decks with a larger wins (despite also with even larger losses). At around 20 actions, ADD performs better than baselines in learning about the decks with the better gains. In all three agent pools (MAB agents, CB agents, RL agents), we observed interesting trajectories revealed by the short-term dynamics (Figure \ref{fig:IGT}), suggesting a promising next step to map from behavioral trajectories to clinically relevant reward processing bias of the human subjects.
 


\subsection{PacMan game across various stationarities} 


We demonstrate the merits of the proposed algorithm using the classic game of PacMan. The goal of the agent is to eat all the dots in the maze, known as Pac-Dots, as soon as possible while simultaneously avoiding collision with ghosts, which roam the maze trying to kill PacMan. The rules for the environment (adopted from Berkeley AI PacMan \footnote{\url{http://ai.berkeley.edu/project_overview.html}}) are as follows. There are two types of negative rewards: on collision with a ghost, PacMan loses the game and gets a negative reward of $-500$; and at each time frame, there is a constant time-penalty of $-1$ for every step taken. There are three types of positive rewards. On eating a Pac-Dot, the agent obtains a reward of $+10$. On successfully eating all the Pac-Dots, the agent wins the game and obtains a reward of $+500$. The game also has two special dots called Power Pellets in the corners of the maze, which on consumption, give PacMan the temporary ability of “eating” ghosts. During this phase, the ghosts are in a ``scared'' state for 40 frames and move at half their speed. On eating a ``scared'' ghost, the agent gets a reward of $+200$, the ghost returns to the center box and returns to its normal ``unscared'' state. As a more realistic scenarios as real-world agents, we define the agents to receive their rewards in positive and negative streams separately. Traditional agents sum the two streams as a regular reward, while Split agents use two streams separately.

We applied several types of stationarities to PacMan as in \cite{lin2020diabolic}. In order to simulate a lifelong learning setting, we assume that the environmental settings arrive in batches (or stages) of episodes, and the specific rule of the game (i.e., reward distributions) may change across batches, while remaining stationary within each batch. The change is defined by a stochastic process of the game setting that an event $A$ is defined for the positive stream and an event $B$ is defined for the negative stream, independent of each other ($A \perp B$). The stochastic process is resampled every 10 rounds (i.e. a batch size of 10). 

\textbf{Stochastic reward muting.} To simulate the changes of turning on or off of a certain reward stream, we define the event $A$ as turning off the positive reward stream (i.e. all the positive rewards are set to be zero) and the event $B$ as turning off the negative reward stream (i.e. all the penalties are set to be zero). $\mathbb{P}(A)=\mathbb{P}(B)=0.5$ in the experiments.

\textbf{Stochastic reward scaling.} To simulate the changes of scaling up a certain reward stream, we define the event $A$ as scaling up the positive reward stream by 100 (i.e. all the positive rewards are multiplied by 100) and the event $B$ as scaling up the negative reward stream (i.e. all the penalties are multiplied by 100). $\mathbb{P}(A)=\mathbb{P}(B)=0.5$ in the experiments.

\textbf{Stochastic reward flipping.} To simulate the changes of flipping certain reward stream, we define the event $A$ as flipping the positive reward stream (i.e. all the positive rewards are multiplied by -1 and considered penalties) and the event $B$ as flipping the negative reward stream (i.e. all the penalties are multiplied by -1 and considered positive rewards). We set $\mathbb{P}(A)=\mathbb{P}(B)=0.5$.

We ran the proposed agents across these different stationarities for 200 episodes over multiple runs and plotted their average final scores with standard errors.

\textbf{Results.} 
 As in Figure \ref{fig:pacman_results}, in all four scenarios, the split models demonstrated competitive performance against their baselines. In the CB agent pools, where the state-less agents were not designed for such a complicated gaming environment, we still observe a converging learning behaviors from these agents. LinUCB as a CB baseline, performed better than the SCTS, which suggested a potentially better theoretical model to integrate split mechanism for this game environment. However, it is worth noting that in the reward flipping scenario, several mental agents are even more advantageous than the standard split models as in Figure \ref{fig:pacman_results}(d), which matches clinical discoveries and the theory of evolutionary psychiatry. For instance, ADHD-like fast-switching attention seems to be especially beneficial in this very non-stationary setting of flipping reward streams. Even in a full stationary setting, the behaviors of these mental agents can have interesting clinical implications. For instance, the video of a CP (``chronic pain'') agent playing PacMan shows a clear avoidance behavior to penalties by staying at a corner very distant from the ghosts and a comparatively lack of interest to reward pursuit by not eating nearby Pac-Dots, matching the clinical characters of chronic pain patients. From the video, we observe that the agent ignored all the rewards in front of it and spent its life hiding from the ghosts, trying to elongate its life span at all costs, even if that implies a constant time penalty to a very negative final score. (The videos of the mental agents playing PacMan after training here\footnote{\url{https://github.com/doerlbh/mentalRL/tree/master/video}})



%% file: sec_tables.tex
 \begin{table*}[ht!]
\begin{minipage}{\linewidth}
      \caption{\textbf{MAB Task:} 100 randomly generated scenarios of Bi-modal rewards}
      \label{tab:MAB} 
      \centering
      \resizebox{0.8\linewidth}{!}{
 \begin{tabular}{ l |x{1.2cm}|x{1.2cm}|x{1.2cm}|x{1.2cm}|x{1.2cm}|x{1.2cm}|x{1.2cm}|x{1.2cm}  }
 &\multicolumn{5}{c}{Baseline} \vline & \multicolumn{3}{c}{Variants of Split MAB agents} \\
 \textbf{MAB} &  TS & UCB1 & EXP3 & gEXP3 & eGreedy & HBTS & PTS & NTS \tabularnewline \hline
TS & - & 31:\textbf{49} & \textbf{71}:9 & \textbf{73}:7 & \textbf{44}:36 & 32:\textbf{48} & \textbf{46}:34 & \textbf{73}:7\tabularnewline
UCB1 & \textbf{49}:31 & - & \textbf{74}:6 & \textbf{77}:3 & \textbf{55}:25 & 34:\textbf{46} & \textbf{54}:26 & \textbf{74}:6\tabularnewline
EXP3 & 9:\textbf{71} & 6:\textbf{74} & - & \textbf{41}:39 & 6:\textbf{74} & 10:\textbf{70} & 12:\textbf{68} & 13:\textbf{67}\tabularnewline
gEXP3 & 7:\textbf{73} & 3:\textbf{77} & 39:\textbf{41} & - & 6:\textbf{74} & 11:\textbf{69} & 10:\textbf{70} & 10:\textbf{70}\tabularnewline
eGreedy & 36:\textbf{44} & 25:\textbf{55} & \textbf{74}:6 & \textbf{74}:6 & - & 28:\textbf{52} & \textbf{48}:32 & \textbf{72}:8\tabularnewline
HBTS & \textbf{48}:32 & \textbf{46}:34 & \textbf{70}:10 & \textbf{69}:11 & \textbf{52}:28 & - & \textbf{59}:21 & \textbf{68}:12\tabularnewline
PTS & 34:\textbf{46} & 26:\textbf{54} & \textbf{68}:12 & \textbf{70}:10 & 32:\textbf{48} & 21:\textbf{59} & - & \textbf{52}:28\tabularnewline
NTS & 7:\textbf{73} & 6:\textbf{74} & \textbf{67}:13 & \textbf{70}:10 & 8:\textbf{72} & 12:\textbf{68} & 28:\textbf{52} & -\tabularnewline
\hline
avg wins (\%)  & 46.72 & \textbf{52.65} & 12.25 & 10.86 & 45.08 & \textbf{52.02} & 38.26 & 25.00 \tabularnewline
 \end{tabular}
 }
    \vspace{0.1in}

      \resizebox{0.8\linewidth}{!}{
 \begin{tabular}{ l | x{1.2cm}|x{1.2cm}|x{1.2cm}|x{1.2cm}| x{1.5cm}|x{1.5cm}  }
 &\multicolumn{3}{c}{Baseline} \vline & \multicolumn{3}{c}{Variants of Split CB Agents} \\
 \textbf{CB} &  CTS & LinUCB & EXP4 & SCTS & PCTS & NCTS \tabularnewline \hline
CTS & - & 19:\textbf{61} & \textbf{73}:7 & \textbf{49}:31 & \textbf{48}:32 & \textbf{67}:13\tabularnewline
LinUCB & \textbf{61}:19 & - & \textbf{76}:4 & \textbf{71}:9 & \textbf{56}:24 & \textbf{75}:5\tabularnewline
EXP4 & 7:\textbf{73} & 4:\textbf{76} & - & 2:\textbf{78} & 7:\textbf{73} & 10:\textbf{70}\tabularnewline
SCTS & 31:\textbf{49} & 9:\textbf{71} & \textbf{78}:2 & - & \textbf{46}:34 & \textbf{71}:9\tabularnewline
PCTS & 32:\textbf{48} & 24:\textbf{56} & \textbf{73}:7 & 34:\textbf{46} & - & \textbf{68}:12\tabularnewline
NCTS & 13:\textbf{67} & 5:\textbf{75} & \textbf{70}:10 & 9:\textbf{71} & 12:\textbf{68} & -\tabularnewline
\hline
avg wins (\%)  & 43.10 & \textbf{57.07} & 5.05 & 39.56 & 38.89 & 18.35\tabularnewline
 \end{tabular}
 }
    \vspace{0.1in}

      \resizebox{0.8\linewidth}{!}{
 \begin{tabular}{ l | c | c | c | c | c | c | c | c }
 &\multicolumn{3}{c}{Baseline} \vline & \multicolumn{5}{c}{Variants of Split RL agents} \\
 \textbf{RL} &  QL & DQL & SARSA & SQL-alg1 & SQL-alg2 & MP & PQL & NQL \\ \hline
QL & - & 39:\textbf{41} & 34:\textbf{46} & \textbf{43}:37 & \textbf{43}:37 & \textbf{42}:38 & \textbf{59}:21 & \textbf{46}:34\\
DQL & \textbf{41}:39 & - & 38:\textbf{42} & \textbf{40}:\textbf{40} & \textbf{44}:36 & \textbf{44}:36 & \textbf{59}:21 & \textbf{46}:34\\
SARSA & \textbf{46}:34 & \textbf{42}:38 & - & \textbf{44}:36 & \textbf{45}:35 & \textbf{44}:36 & \textbf{51}:29 & \textbf{48}:32\\
SQL & 37:\textbf{43} & \textbf{40}:\textbf{40} & 36:\textbf{44} & - & \textbf{41}:39 & 38:\textbf{42} & \textbf{59}:21 & \textbf{46}:34\\
SQL2 & 37:\textbf{43} & 36:\textbf{44} & 35:\textbf{45} & 39:\textbf{41} & - & \textbf{42}:38 & \textbf{55}:25 & \textbf{48}:32\\
MP & 38:\textbf{42} & 36:\textbf{44} & 36:\textbf{44} & \textbf{42}:38 & 38:\textbf{42} & - & \textbf{52}:28 & \textbf{42}:38\\
PQL & 21:\textbf{59} & 21:\textbf{59} & 29:\textbf{51} & 21:\textbf{59} & 25:\textbf{55} & 28:\textbf{52} & - & 32:\textbf{48}\\
NQL & 34:\textbf{46} & 34:\textbf{46} & 32:\textbf{48} & 34:\textbf{46} & 32:\textbf{48} & 38:\textbf{42} & \textbf{48}:32 & -\\
\hline
avg wins (\%)  & 38.64 & 39.39 &\textbf{ 40.40} & 37.50 & 36.87 & 35.86 & 22.35 & 31.82\\
 \end{tabular}
 }
 \end{minipage}
     \vspace{0.1in}

  \centering
        \caption{\textbf{``Mental'' Agents in MAB Task:} 100 randomly generated scenarios}
      \label{tab:MABmental} 
 \begin{minipage}{.8\linewidth}
      \centering
      \resizebox{1\linewidth}{!}{
 \begin{tabular}{  l | c | c | c | c | c| c | c | c }
 \textbf{MAB}  & b-ADD & b-ADHD  & b-AD  & b-CP & b-bvFTD & b-PD & b-M &  avg wins (\%)   \\ \hline
TS & 39:\textbf{41} & 38:\textbf{42} & 39:\textbf{41} & \textbf{41}:39 & 39:\textbf{41} & 33:\textbf{47} & 30:\textbf{50} & 37.37\\
UCB1 & \textbf{50}:30 & \textbf{43}:37 & \textbf{54}:26 & \textbf{45}:35 & \textbf{52}:28 & 38:\textbf{42} & \textbf{42}:38 & 46.75\\
EXP3 & 6:\textbf{74} & 12:\textbf{68} & 7:\textbf{73} & 8:\textbf{72} & 7:\textbf{73} & 9:\textbf{71} & 6:\textbf{74} & 7.94\\
eGreedy & \textbf{43}:37 & 32:\textbf{48} & 36:\textbf{44} & 38:\textbf{42} & 37:\textbf{43} & 34:\textbf{46} & 30:\textbf{50} & 36.08\\
HBTS & \textbf{52}:28 & \textbf{40}:\textbf{40} & \textbf{45}:35 & \textbf{51}:29 & \textbf{47}:33 & 38:\textbf{42} & 38:\textbf{42} & 44.88\\
\hline
avg wins (\%)  & 42.42 & 47.47 & 44.24 & 43.84 & 44.04 & 50.10 & 51.31
 \end{tabular}
 }
 
 \vspace{0.1in}
       \resizebox{1\linewidth}{!}{
 \begin{tabular}{  l | c | c | c | c | c| c | c | c }
 \textbf{CB}  & cb-ADD & cb-ADHD  & cb-AD  & cb-CP & cb-bvFTD & cb-PD & cb-M &  avg wins (\%)   \\ \hline
CTS & \textbf{68}:12 & \textbf{47}:33 & \textbf{72}:8 & \textbf{40}:\textbf{40} & \textbf{67}:13 & \textbf{68}:12 & \textbf{61}:19 & 61.04\\
LinUCB & \textbf{75}:5 & \textbf{56}:24 & \textbf{77}:3 & \textbf{53}:27 & \textbf{74}:6 & \textbf{76}:4 & \textbf{72}:8 & 69.70\\
EXP4 & 21:\textbf{59} & 5:\textbf{75} & 18:\textbf{62} & 9:\textbf{71} & 9:\textbf{71} & 10:\textbf{70} & 15:\textbf{65} & 12.55\\
SCTS & \textbf{73}:7 & 39:\textbf{41} & \textbf{74}:6 & 36:\textbf{44} & \textbf{70}:10 & \textbf{73}:7 & \textbf{65}:15 & 62.05\\
\hline
avg wins (\%)  & 20.96 & 43.69 & 19.95 & 45.96 & 25.25 & 23.48 & 27.02
 \end{tabular}
 }
 
  \vspace{0.1in}
\resizebox{1\linewidth}{!}{
 \begin{tabular}{  l | c | c | c | c | c| c | c | c }
 \textbf{RL}  & ADD & ADHD  & AD  & CP & bvFTD & PD & M &  avg wins (\%)   \\ \hline
QL & \textbf{65}:15 & \textbf{59}:21 & \textbf{55}:25 & \textbf{64}:16 & \textbf{54}:26 & \textbf{59}:21 & \textbf{56}:24 & 59.45\\
DQL & \textbf{62}:18 & \textbf{62}:18 & \textbf{58}:22 & \textbf{62}:18 & \textbf{49}:31 & \textbf{56}:24 & \textbf{50}:30 & 57.58\\
SARSA & \textbf{57}:23 & \textbf{57}:23 & \textbf{59}:21 & \textbf{63}:17 & \textbf{51}:29 & \textbf{59}:21 & \textbf{53}:27 & 57.58\\
SQL & \textbf{57}:23 & \textbf{54}:26 & \textbf{48}:32 & \textbf{61}:19 & \textbf{50}:30 & \textbf{52}:28 & \textbf{50}:30 & 53.68\\
\hline
avg wins (\%)  & 19.95 & 22.22 & 25.25 & 17.68 & 29.29 & 23.74 & 28.03
 \end{tabular}
 }
 \end{minipage}
 \end{table*}

\begin{table*}[ht!]
\begin{minipage}{\linewidth}
      \caption{\textbf{MDP Task:} 100 randomly generated scenarios of Bi-modal rewards}
      \label{tab:MDP} 
      \centering
      \resizebox{0.8\linewidth}{!}{
 \begin{tabular}{ l |x{1.2cm}|x{1.2cm}|x{1.2cm}|x{1.2cm}|x{1.2cm}|x{1.2cm}|x{1.2cm}|x{1.2cm}  }
 &\multicolumn{5}{c}{Baseline} \vline & \multicolumn{3}{c}{Variants of Split MAB agents} \\
 \textbf{MAB} &  TS & UCB1 & EXP3 & gEXP3 & eGreedy & HBTS & PTS & NTS \tabularnewline \hline
TS & - & \textbf{42}:38 & 38:\textbf{42} & 37:\textbf{43} & \textbf{43}:37 & \textbf{40}:\textbf{40} & \textbf{49}:31 & \textbf{44}:36\tabularnewline
UCB1 & 38:\textbf{42} & - & 39:\textbf{41} & 29:\textbf{51} & \textbf{44}:36 & 33:\textbf{47} & \textbf{42}:38 & \textbf{43}:37\tabularnewline
EXP3 & \textbf{42}:38 & \textbf{41}:39 & - & 35:\textbf{45} & 39:\textbf{41} & \textbf{43}:37 & \textbf{45}:35 & \textbf{46}:34\tabularnewline
gEXP3 & \textbf{43}:37 & \textbf{51}:29 & \textbf{45}:35 & - & \textbf{42}:38 & \textbf{43}:37 & \textbf{45}:35 & \textbf{47}:33\tabularnewline
eGreedy & 37:\textbf{43} & 36:\textbf{44} & \textbf{41}:39 & 38:\textbf{42} & - & 38:\textbf{42} & 38:\textbf{42} & 36:\textbf{44}\tabularnewline
HBTS & \textbf{40}:\textbf{40} & \textbf{47}:33 & 37:\textbf{43} & 37:\textbf{43} & \textbf{42}:38 & - & 39:\textbf{41} & \textbf{48}:32\tabularnewline
PTS & 31:\textbf{49} & 38:\textbf{42} & 35:\textbf{45} & 35:\textbf{45} & \textbf{42}:38 & \textbf{41}:39 & - & 37:\textbf{43}\tabularnewline
NTS & 36:\textbf{44} & 37:\textbf{43} & 34:\textbf{46} & 33:\textbf{47} & \textbf{44}:36 & 32:\textbf{48} & \textbf{43}:37 & -\tabularnewline
\hline
avg wins (\%)  & 36.99 & 33.84 & 36.74 & \textbf{39.90} & 33.33 & 36.62 & 32.70 & 32.70\tabularnewline
 \end{tabular}
 }
    \vspace{0.1in}

      \resizebox{0.8\linewidth}{!}{
 \begin{tabular}{ l | x{1.2cm}|x{1.2cm}|x{1.2cm}|x{1.2cm}| x{1.5cm}|x{1.5cm}  }
 &\multicolumn{3}{c}{Baseline} \vline & \multicolumn{3}{c}{Variants of Split CB Agents} \\
 \textbf{CB} &  CTS & LinUCB & EXP4 & SCTS & PCTS & NCTS \tabularnewline \hline
CTS & - & 6:\textbf{74} & 36:\textbf{44} & \textbf{42}:38 & 30:\textbf{50} & 37:\textbf{43}\tabularnewline
LinUCB & \textbf{74}:6 & - & \textbf{74}:6 & \textbf{74}:6 & \textbf{72}:8 & \textbf{75}:5\tabularnewline
EXP4 & \textbf{44}:36 & 6:\textbf{74} & - & \textbf{45}:35 & 31:\textbf{49} & \textbf{41}:39\tabularnewline
SCTS & 38:\textbf{42} & 6:\textbf{74} & 35:\textbf{45} & - & 30:\textbf{50} & 39:\textbf{41}\tabularnewline
PCTS & \textbf{50}:30 & 8:\textbf{72} & \textbf{49}:31 & \textbf{50}:30 & - & \textbf{50}:30\tabularnewline
NCTS & \textbf{43}:37 & 5:\textbf{75} & 39:\textbf{41} & \textbf{41}:39 & 30:\textbf{50} & -\tabularnewline
\hline
avg wins (\%)  & 25.42 & \textbf{62.12} & 28.11 & 24.92 & 34.85 & 26.60\tabularnewline
 \end{tabular}
 }
    \vspace{0.1in}

      \resizebox{0.8\linewidth}{!}{
 \begin{tabular}{ l | c | c | c | c | c | c | c | c }
 &\multicolumn{3}{c}{Baseline} \vline & \multicolumn{5}{c}{Variants of Split RL agents} \\
 \textbf{RL} &  QL & DQL & SARSA & SQL-alg1 & SQL-alg2 & MP & PQL & NQL \\ \hline
QL & - & \textbf{62}:38 & \textbf{55}:45 & \textbf{63}:37 & \textbf{54}:46 & 47:\textbf{53} & \textbf{65}:35 & \textbf{90}:10\\
DQL & 38:\textbf{62} & - & 40:\textbf{60} & 48:\textbf{52} & 48:\textbf{52} & 43:\textbf{57} & \textbf{55}:45 & \textbf{86}:14\\
SARSA & 45:\textbf{55} & \textbf{60}:40 & - & \textbf{63}:37 & \textbf{51}:49 & \textbf{52}:48 & \textbf{64}:36 & \textbf{88}:12\\
SQL & 37:\textbf{63} & \textbf{52}:48 & 37:\textbf{63} & - & 42:\textbf{58} & 26:\textbf{74} & \textbf{55}:45 & \textbf{72}:28\\
SQL2 & 46:\textbf{54} & \textbf{52}:48 & 49:\textbf{51} & \textbf{58}:42 & - & 39:\textbf{61} & \textbf{64}:36 & \textbf{72}:28\\
MP & \textbf{53}:47 & \textbf{57}:43 & 48:\textbf{52} & \textbf{74}:26 & \textbf{61}:39 & - & \textbf{66}:34 & \textbf{82}:18\\
PQL & 35:\textbf{65} & 45:\textbf{55} & 36:\textbf{64} & 45:\textbf{55} & 36:\textbf{64} & 34:\textbf{66} & - & \textbf{68}:32\\
NQL & 10:\textbf{90} & 14:\textbf{86} & 12:\textbf{88} & 28:\textbf{72} & 28:\textbf{72} & 18:\textbf{82} & 32:\textbf{68} & -\\
\hline
avg wins (\%)  & 55.05 & 45.20 & 53.41 & 40.53 & 47.98 & \textbf{55.68} & 37.75 & 17.93\\
 \end{tabular}
 }
 \end{minipage}
     \vspace{0.1in}

  \centering
        \caption{\textbf{``Mental'' Agents in MDP Task:} 100 randomly generated scenarios}
      \label{tab:MDPmental} 
 \begin{minipage}{.8\linewidth}
      \centering
      \resizebox{1\linewidth}{!}{
 \begin{tabular}{  l | c | c | c | c | c| c | c | c }
 \textbf{MAB}  & b-ADD & b-ADHD  & b-AD  & b-CP & b-bvFTD & b-PD & b-M &  avg wins (\%)   \\ \hline
TS & \textbf{43}:37 & \textbf{49}:31 & \textbf{45}:35 & \textbf{45}:35 & \textbf{44}:36 & 39:\textbf{41} & 36:\textbf{44} & 43.43\\
UCB1 & 38:\textbf{42} & \textbf{48}:32 & \textbf{41}:39 & \textbf{40}:\textbf{40} & 39:\textbf{41} & 39:\textbf{41} & 36:\textbf{44} & 40.55\\
EXP3 & 38:\textbf{42} & \textbf{47}:33 & \textbf{46}:34 & \textbf{41}:39 & \textbf{41}:39 & \textbf{40}:\textbf{40} & 36:\textbf{44} & 41.70\\
eGreedy & \textbf{40}:\textbf{40} & \textbf{44}:36 & \textbf{41}:39 & 38:\textbf{42} & \textbf{41}:39 & 35:\textbf{45} & 39:\textbf{41} & 40.12\\
HBTS & \textbf{40}:\textbf{40} & \textbf{48}:32 & \textbf{47}:33 & \textbf{43}:37 & \textbf{49}:31 & \textbf{42}:38 & 39:\textbf{41} & 44.44\\
\hline
avg wins (\%)  & 40.61 & 33.13 & 36.36 & 38.99 & 37.58 & 41.41 & 43.23
 \end{tabular}
 }
 \vspace{0.1in}
 
       \resizebox{1\linewidth}{!}{
 \begin{tabular}{  l | c | c | c | c | c| c | c | c }
 \textbf{CB}  & cb-ADD & cb-ADHD  & cb-AD  & cb-CP & cb-bvFTD & cb-PD & cb-M &  avg wins (\%)   \\ \hline
CTS & 37:\textbf{43} & \textbf{41}:39 & 36:\textbf{44} & 35:\textbf{45} & 35:\textbf{45} & 38:\textbf{42} & \textbf{45}:35 & 38.53\\
LinUCB & \textbf{73}:7 & \textbf{76}:4 & \textbf{73}:7 & \textbf{74}:6 & \textbf{75}:5 & \textbf{74}:6 & \textbf{76}:4 & 75.18\\
EXP4 & 38:\textbf{42} & 38:\textbf{42} & 33:\textbf{47} & \textbf{42}:38 & \textbf{41}:39 & \textbf{44}:36 & \textbf{44}:36 & 40.40\\
SCTS & 36:\textbf{44} & \textbf{41}:39 & 31:\textbf{49} & 39:\textbf{41} & 37:\textbf{43} & \textbf{40}:\textbf{40} & \textbf{43}:37 & 38.53\\
\hline
avg wins (\%)  & 34.34 & 31.31 & 37.12 & 32.83 & 33.33 & 31.31 & 28.28
 \end{tabular}
 }
  \vspace{0.1in}
  
\resizebox{1\linewidth}{!}{
 \begin{tabular}{  l | c | c | c | c | c| c | c | c }
 \textbf{RL}  & ADD & ADHD  & AD  & CP & bvFTD & PD & M &  avg wins (\%)   \\ \hline
QL & \textbf{70}:10 & \textbf{44}:36 & \textbf{67}:13 & \textbf{48}:32 & \textbf{58}:22 & \textbf{49}:31 & \textbf{48}:32 & 55.41\\
DQL & \textbf{69}:11 & \textbf{42}:38 & \textbf{66}:14 & \textbf{45}:35 & \textbf{56}:24 & \textbf{45}:35 & \textbf{56}:24 & 54.69\\
SARSA & \textbf{75}:5 & \textbf{48}:32 & \textbf{71}:9 & \textbf{53}:27 & \textbf{61}:19 & \textbf{52}:28 & \textbf{54}:26 & 59.74\\
SQL & \textbf{68}:12 & \textbf{41}:39 & \textbf{60}:20 & 38:\textbf{42} & \textbf{54}:26 & \textbf{44}:36 & \textbf{43}:37 & 50.22\\
\hline
avg wins (\%)  & 9.60 & 36.62 & 14.14 & 34.34 & 22.98 & 32.83 & 30.05
 \end{tabular}
 }
 \end{minipage}
 \end{table*}

%% file: sec_conclusion.tex
\section{Conclusions}
\label{sec:conclusion}


This research proposes a novel parametric family of algorithms for multi-armed bandits, contextual bandits and RL problems, extending the classical algorithms to model a wide range of potential reward processing biases. Our approach draws an inspiration from extensive literature on decision-making behavior in neurological and psychiatric disorders stemming from disturbances of the reward processing system, and demonstrates high flexibility of our multi-parameter model which allows to tune the weights on incoming two-stream rewards and memories about the prior reward history. Our preliminary results support multiple prior observations about reward processing biases in a range of mental disorders, thus indicating the potential of the proposed model and its future extensions to capture reward-processing aspects across various neurological and psychiatric conditions. 

The contribution of this research is two-fold: from the machine learning perspective, we propose a simple yet powerful and more adaptive approach to MAB, CB and RL problems; from the neuroscience perspective, this work is the first attempt at a general, unifying model of reward processing and its disruptions across a wide population including both healthy subjects and those with mental disorders, which has a potential to become a useful computational tool for neuroscientists and psychiatrists studying such disorders. Among the directions for future work, we plan to investigate the optimal parameters in a series of computer games evaluated on different criteria, for example, longest survival time vs. highest final score. Further work includes exploring the multi-agent interactions given different reward processing bias. These discoveries can help build more interpretable real-world humanoid decision making systems. On the neuroscience side, the next steps would include further tuning and extending the proposed model to better capture observations in modern literature, as well as testing the model on both healthy subjects and patients with mental conditions.

%% file: sec_appendix.tex
\appendix

\section{Further Motivation from Neuroscience}
\label{sec:neuro}

In the following section, we provide further discussion with a literature review on the neuroscience and clinical studies related to the reward processing systems.

\textbf{Cellular computation of reward and reward violation.} Decades of evidence has linked dopamine function to reinforcement learning via neurons in the midbrain and its connections in the basal ganglia, limbic regions, and cortex. Firing rates of dopamine neurons computationally represent reward magnitude, expectancy, and violations (prediction error) and other value-based signals \cite{Schultz1997}. This allows an animal to update and maintain value expectations associated with particular states and actions. When functioning properly, this helps an animal develop a policy to maximize outcomes by approaching/choosing cues with higher expected value and avoiding cues associated with loss or punishment. The mechanism is conceptually similar to  reinforcement learning widely used in computing and robotics \cite{Sutton1998}, suggesting mechanistic overlap in humans and AI. Evidence of Q-learning and actor-critic models have been observed in spiking activity in midbrain dopamine neurons in primates \cite{Bayer2005} and in the human striatum using the BOLD signal \cite{ODoherty2004}. 

\textbf{Positive vs. negative learning signals.} Phasic dopamine signaling represents bidirectional (positive and negative) coding for prediction error signals \cite{Hart2014}, but underlying mechanisms show differentiation for reward relative to punishment learning \cite{Seymour2007}. Though representation of cellular-level aversive error signaling has been debated \cite{Dayan2008}, it is widely thought that rewarding, salient information is represented by phasic dopamine signals, whereas reward omission or punishment signals are represented by dips or pauses in baseline dopamine firing \cite{Schultz1997}. These mechanisms have downstream effects on motivation, approach behavior, and action selection. Reward signaling in a direct pathway links striatum to cortex via dopamine neurons that disinhibit the thalamus via the internal segment of the globus pallidus and facilitate action and approach behavior. Alternatively, aversive signals may have an opposite effect in the indirect pathway mediated by D2 neurons inhibiting thalamic function and ultimately action, as well \cite{Frank2006}. Manipulating these circuits through pharmacological measures or disease has demonstrated computationally-predictable effects that bias learning from positive or negative prediction error in humans \cite{frank2004carrot}, and contribute to our understanding of perceptible differences in human decision making when differentially motivated by loss or gain \cite{Tversky1981}.

\textbf{Clinical Implications.} Highlighting the importance of using computational models to understand predict disease outcomes, many symptoms of neurological and psychiatric disease are related to biases in learning from positive and negative feedback \cite{Maia2011}. Studies in humans have shown that when reward signaling in the direct pathway is over-expressed, this may enhance the value associated with a state and incur pathological reward-seeking behavior, like gambling or substance use. Conversely, when aversive error signals are enhanced, this results in dampening of reward experience and increased motor inhibition, causing symptoms that decrease motivation, such as apathy, social withdrawal, fatigue, and depression. Further, it has been proposed that exposure to a particular distribution of experiences during critical periods of development can biologically predispose an individual to learn from positive or negative outcomes, making them more or less susceptible to risk for brain-based illnesses \cite{Holmes2018}. These points distinctly highlight the need for a greater understanding of how intelligent systems differentially learn from rewards or punishments, and how experience sampling may impact reinforcement learning during influential training periods.